\pdfoutput=1

\documentclass[11pt]{article}

\usepackage{ACL2023}

\usepackage{times}
\usepackage{latexsym}
\usepackage{enumerate}
\usepackage{enumitem}

\usepackage[T1]{fontenc}

\usepackage[utf8]{inputenc}

\usepackage{microtype}

\usepackage{inconsolata}
\usepackage{graphicx}
\usepackage{multirow}
\usepackage{xcolor}
\usepackage{colortbl}
\usepackage{booktabs}
\usepackage{ulem}
\usepackage{array}
\usepackage{amsfonts,amssymb}
\usepackage{amsmath}
\usepackage[ruled,linesnumbered,noend]{algorithm2e}
\usepackage{algpseudocode}
\usepackage{subfigure}
\usepackage{arydshln}
\usepackage{pifont}

\title{LMPT: Prompt Tuning with Class-Specific Embedding Loss for Long-Tailed Multi-Label Visual Recognition}

\author{Peng Xia$^{1}$, Di Xu$^{2}$, Ming Hu$^1$, \textbf{Lie Ju$^1$, Zongyuan Ge$^1$}\\ $^1$Monash University, $^2$Imperial College London\\ \texttt{richard.peng.xia@gmail.com, zongyuan.ge@monash.edu}}

\begin{document}
\maketitle
\begin{abstract}
Long-tailed multi-label visual recognition (LTML) task is a highly challenging task due to the label co-occurrence and imbalanced data distribution. In this work, we propose a unified framework for LTML, namely prompt tuning with class-specific embedding loss (LMPT), capturing the semantic feature interactions between categories by combining text and image modality data and improving the performance synchronously on both head and tail classes. Specifically, LMPT introduces the embedding loss function with class-aware soft margin and re-weighting to learn class-specific contexts with the benefit of textual descriptions (captions), which could help establish semantic relationships between classes, especially between the head and tail classes. Furthermore, taking into account the class imbalance, the distribution-balanced loss is adopted as the classification loss function to further improve the performance on the tail classes without compromising head classes. Extensive experiments are conducted on VOC-LT and COCO-LT datasets, which demonstrates that our method significantly surpasses the previous state-of-the-art methods and zero-shot CLIP in LTML. Our codes are fully public at \url{https://github.com/richard-peng-xia/LMPT}.
\end{abstract}

\section{Introduction}

\begin{figure}[htbp]
    \centering
        \begin{minipage}{0.45\linewidth}
        \centerline{\includegraphics[width=0.9\textwidth]{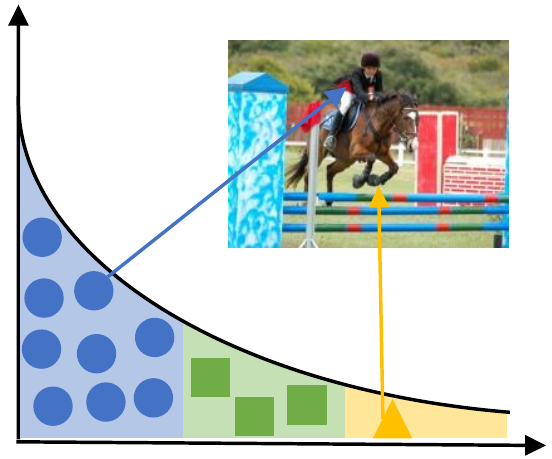}}
        \centerline{(a)}
        \label{fig:fig1a}
	\end{minipage}
 \medskip
	\begin{minipage}{0.45\linewidth}
	\centerline{\includegraphics[width=0.9\textwidth]{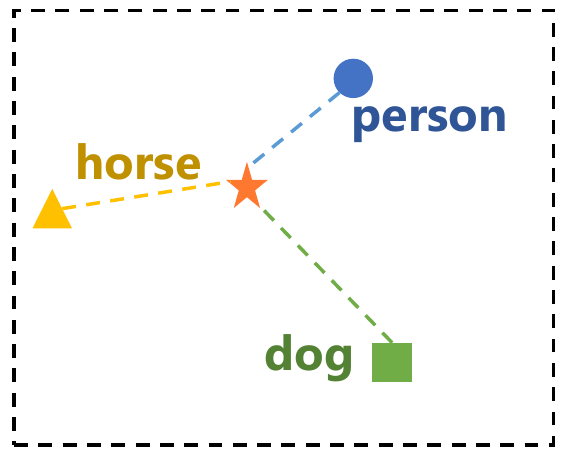}}
	\centerline{(b)}
        \label{fig:fig1b}
	\end{minipage}

        \begin{minipage}{0.45\linewidth}
        \centerline{\includegraphics[width=0.9\textwidth]{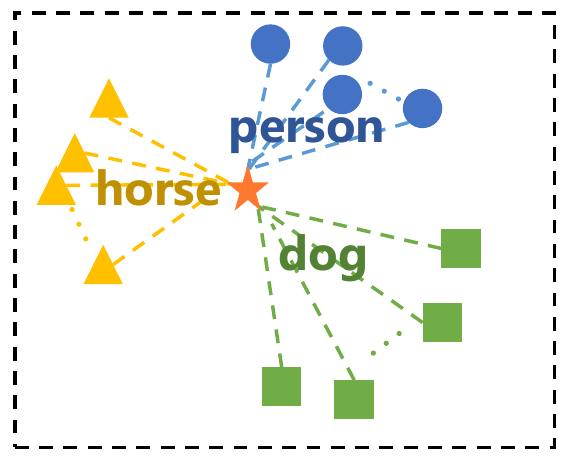}}
		\centerline{(c)}
        \label{fig:fig1c}
        \end{minipage}
\medskip
        \begin{minipage}{0.45\linewidth}
        \centerline{\includegraphics[width=0.9\textwidth]{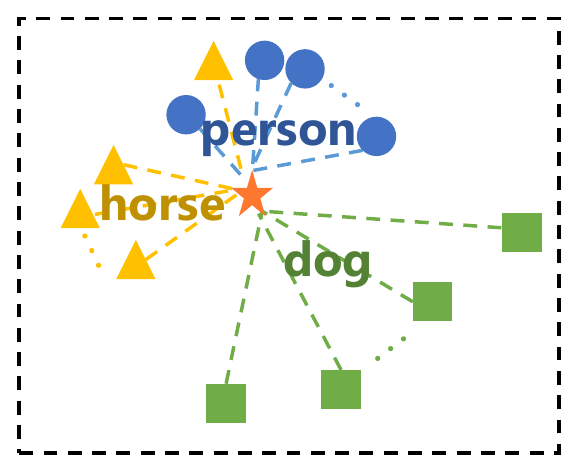}}
		\centerline{(d)}
        \label{fig:fig1d}
        \end{minipage}
\vspace{-0.2in}
\caption{The class distribution is long-tailed and the VLM compares image embeddings\textcolor{red}{$\star$} to text embeddings\textcolor[RGB]{44, 73, 200}{$\bullet$}\textcolor[RGB]{112, 173, 71}{$\blacksquare$}\textcolor[RGB]{255, 204, 0}{$\blacktriangle$} of the class, which means the closer the distance between the embeddings of different modalities, the higher the probability that the category of the text embeddings matches the image. (a) Person and horse in the image belong to the head classes and the tail classes respectively. (b) Zero-Shot CLIP. (c) Exsiting Prompt Tuning \textit{w/o} CSE loss. (d) LMPT (Ours) \textit{w/} CSE loss.}
   \vspace{-0.3in}
\label{fig:motivation}
\end{figure}

\quad Long-tailed multi-label visual recognition (LTML) ~\cite{wu2020distribution,guo2021long} is a common and practical task owing to the highly imbalanced data distribution ~\cite{zhang2021deep} and diverse objects of real-world images~\cite{wang2017multi,ju2021long}. Compared with long-tailed recognition and multi-label recognition tasks, LTML is more complex and challenging, because it requires capturing multiple categories and the label co-occurrence in individual images ~\cite{chen2019learning}, which needs to compensate for the negative impacts caused by the long-tailed distribution (i.e., \textit{low performance on the tail classes}). \par
Several approaches have been proposed to address the LTML problem from different perspectives, such as re-sampling ~\cite{buda2018systematic,dong2017class,guo2021long}, re-weighting ~\cite{cao2019learning,wu2020distribution} and modeling more powerful structures ~\cite{chen2019learning,wang2016cnn,wang2017multi}. Despite their great contributions, these works neglect to take into account two crucial aspects. \textit{First of all}, the importance of semantic feature interaction between classes to capture label co-occurrence. However, these methods are limited to balancing the distribution of categories from the perspective of samples, without considering the feature correlation between different classes. \textit{Second}, synchronous improvements in head-to-tail category performance, while some of these works improve the performance of tail classes at the expense of the head classes. \par
Recently, graphic models have been introduced to model the semantic label correlation in a few works \cite{chen2019learning, wang2016cnn}, whereas these works are complex and are modeling label dependencies mainly based on the image modality without additional semantic information from other modal data. Vision-language models (VLMs)~\cite{radford2021learning,jia2021scaling,tian2022vl,huang2022idea,xia2024cares} demonstrate the huge potential of text modality on semantic context feature for downstream visual tasks, especially for the prompt tuning methods~\cite{schick2021exploiting,shin2020autoprompt,yao2021cpt,xia2023hgclip}, which provide an efficient way to transfer pre-trained VLMs to downstream tasks by learning the task-specific prompts rather than finetuning the entire model. Nonetheless, the existing prompt tuning methods~\cite{zhou2022learning,zhou2022conditional,sun2022dualcoop} for visual recognition simply minimize prediction errors using the classification loss (\textit{e.g.}, cross-entropy loss) with respect to the learnable prompts, which may lead to learning general embeddings or inaccurate class-related embeddings. For instance, when presented with an image (Fig.1a) that contains both a head class $\left[\mathrm{person}\right]$ and a tail class $\left[\mathrm{horse}\right]$, the zero-shot method (Fig.1b) relies solely on the rich knowledge of the pre-trained VLMs to assess the similarity between the image and the word embeddings of the class names, while the existing prompt tuning method (Fig.1c) further learns more generalized prompt tokens to improve model performance. However, these methods do not consider the inter-class relationships, particularly between head and tail classes, which is a critical factor for LTML. This underscores the need for approaches that incorporate such relationships to improve performance in such scenarios.
\par
Therefore, to address these issues, we present the class-specific embedding loss for \textbf{p}rompt \textbf{t}uning on \textbf{l}ong-tailed \textbf{m}ulti-label visual recognition, called \textsc{LMPT}. The abundance of image-caption data facilitates prompt learning that encompasses more nuanced and specific textual descriptions, as well as the semantic inter-dependencies between categories (Fig.1d) that share information, such as similar features or common descriptions. This attribute is particularly critical in the identification of both head and tail classes. More specifically, we propose the class-specific embedding loss to enhance the inclusivity of class-related embeddings within prompts. By gradually approaching the embeddings of the corresponding caption, our proposed approach enables prompt tokens to effectively judge the association between different classes with the aid of textual modality. 
Aiming for class imbalance and consistency improvements between head classes and tail classes, we integrate class-aware soft margin and re-weighting into the class-specific embedding loss, which serves to assign larger margins and more weights to tail classes. Notably, for images containing both head and tail classes, our approach outperforms visual models and current prompt tuning methods.
Moreover, we adopt the distribution-balanced loss~\cite{wu2020distribution} as the classification loss. To sum up, the main contributions of this work include:
\begin{itemize}[leftmargin=*]
    \item  We propose the LMPT framework to adapt pre-trained VLMs to tackle long-tailed multi-label visual recognition, where captions are easily accessible from public image-caption datasets or generated by powerful image-caption models~\cite{wang2022unifying}.
    \item We present a novel class-specific embedding loss with class-aware soft margin and re-weighting to learn more fine-grained and class-related embeddings that build semantic relationships across head and tail classes with shared semantic information. Such design can benefit performance in tail classes and hard-to-recognize classes with the help of text modality. 
    \item We verify the effectiveness of the proposed method by achieving new state-of-the-art (SOTA) results on two datasets, which outperform previous SOTA~\cite{guo2021long} by 9/6\% and zero-shot CLIP by 6/2\% on VOC-LT / COCO-LT. 
\end{itemize}

\section{Related Work}

\begin{figure*}
\begin{center}
\includegraphics[height=6cm,width=15cm]{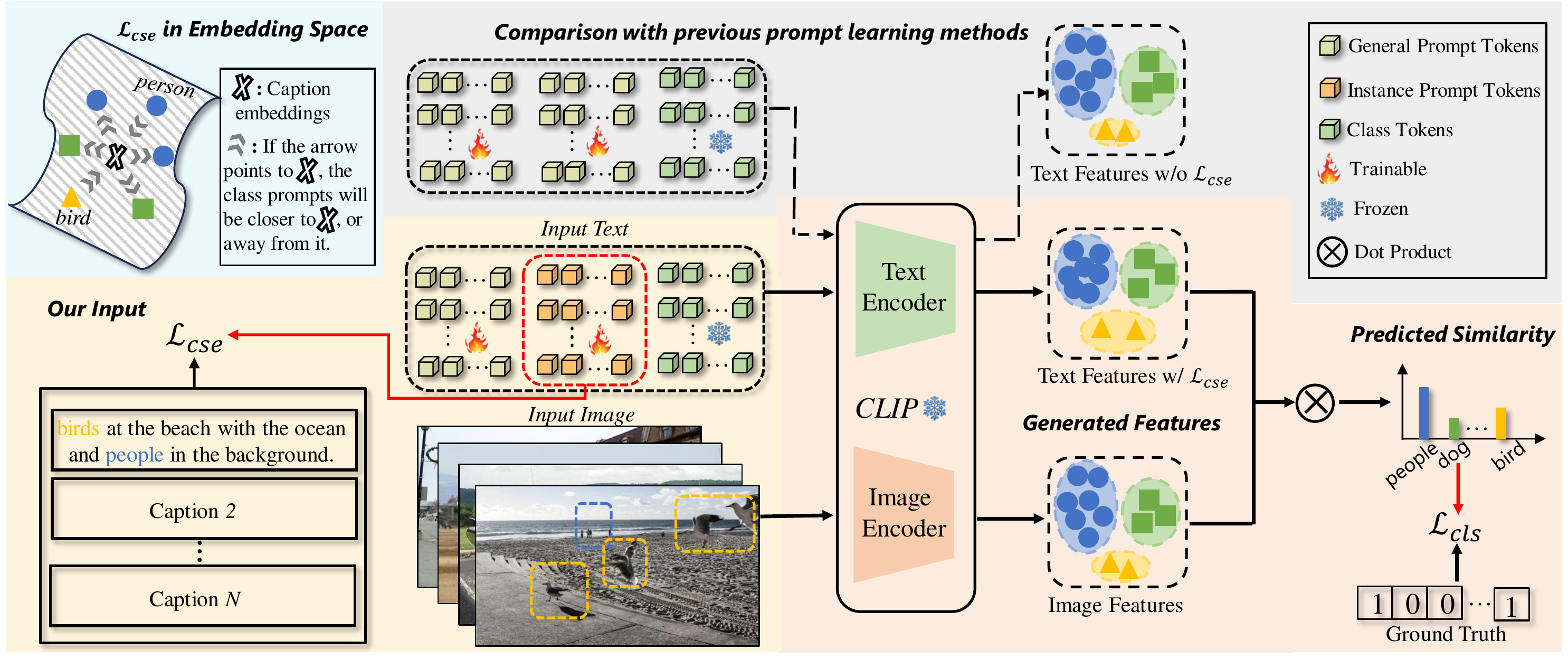}
\end{center}
\caption{Overview of the architecture of our proposed method. The color blocks are defined as shown in Fig. 1.}
\label{fig:pipeline}
\vspace{-0.2em}
\end{figure*}

\subsection{Long-Tailed Visual Recognition}
Real-world training data usually exhibits long-tailed distribution~\cite{zhang2021deep}, which presents a challenge for traditional methods due to the imbalanced class distribution. To address this problem, several approaches~\cite{cui2022reslt, menon2020long, ouyang2016factors, samuel2021distributional} have been proposed from different aspects. One common method is to directly re-sample the training data to balance the class distribution~\cite{drummond2003c4, buda2018systematic, dong2017class}, by adjusting the sampling rate of head classes and tail classes, yet it might lead to the overfitting of tail classes. A better solution is to design re-weighted loss functions~\cite{khan2017cost, huang2016learning, cao2019learning} that assign more weight to tail classes or ignore negative gradients~\cite{tan2020equalization} for tail classes.
In addition, researchers also propose to use techniques such as transfer learning~\cite{liu2019large, zhu2020inflated} and self-supervised learning~\cite{kang2020exploring, zhang2021test} to alleviate the class imbalance problem. Recently, some studies~\cite{ma2021simple, tian2022vl} also explore the possibility of text modality by refining visual-language representations on the long-tailed recognition tasks.

\subsection{Multi-Label Visual Recognition}
For multi-label visual recognition, some early methods include treating it as multiple binary image classifications~\cite{tsoumakas2007multi,zhang2013review} or finding k-nearest neighbors~\cite{zhang2007ml}. To locate regions of interest, some researchers~\cite{wang2016cnn, wang2017multi} proposed to introduce recurrent neural networks (\textit{e.g.}, RNN, LSTM) to learn a joint image-label embedding. In addition, Chen \textit{et al.}~\cite{chen2019learning} proposed to model the label correlations by constructing a graph based on the label co-occurrence and Ye \textit{et al.}~\cite{ye2020attention} updated static graph to dynamic graph convolutional network (GCN) for robust representation. Wu \textit{et al.}~\cite{wu2020distribution} proposed a distribution-balanced loss and Guo \textit{et al.}~\cite{guo2021long} adopted collaborative training on the uniform and re-balanced samplings to alleviate the class imbalanced problem. There is also a popular trend to align between visual and textual features~\cite{xu2022dual,liu2021query2label, huang2022idea, ridnik2023ml} for multi-label recognition.

\subsection{Prompt Tuning for Vision-Language Models}
Prompt tuning~\cite{schick2021exploiting,shin2020autoprompt,yao2021cpt} is a parameter-efficient technique used to utilize the representation ability of pre-trained vision-language models to achieve better performance instead of fine-tuning the whole model on downstream tasks. Meanwhile, large-scale vision-language models (\textit{e.g.}, CLIP~\cite{radford2021learning}, ALIGN~\cite{jia2021scaling}) have demonstrated impressive power to learn visual and textual features. CoOp~\cite{zhou2022learning} learns soft prompts via minimizing the classification loss and CoCoOp~\cite{zhou2022conditional} further formulates the prompts in an image-conditional way to improve its generalization to unseen classes. DualCoOp~\cite{sun2022dualcoop} firstly adapts CLIP to multi-label image recognition by learning pairs of positive and negative prompts for each class, then TaI-DPT~\cite{guo2023texts} extracts both coarse-grained and fine-grained embedding by treating texts as images in prompt tuning. Different from the above work, LMPT focuses on exploring the transfer ability to address long-tailed multi-label visual recognition.

\section{Methodology}

\quad In this section, we present our proposed prompting tuning method, \textit{i.e.}, LMPT, for adapting pre-trained vision-language models for long-tailed multi-label visual recognition. \par 

\subsection{Preliminaries}
Consider $\mathcal{D}$ as the dataset we use, ${N}$ as the number of the dataset, $C$ as the number of classes, and $L$ as the fixed length of contexts for optimization. Then $(x^{k},y^{k},t^{k})\in \mathcal{D}_{train}$, ${k}\in \left\{1, ..., N \right\}$, where $x^{k}$ is an input single image, $y^{k}=\left[y_{1}^{k},...,y_{C}^{k} \right]\in {\left\{0,1 \right\}}^{C}$ is the multi-label ground-truth and $t^{k}=\left[t_{1}^{k}, ..., t_{L}^{k}\right]$ is the corresponding text embedding of text description (caption). But during the test phase, only $(x^{k},y^{k})\in \mathcal{D}_{test}$. Let $n_{i}=\sum_{k=1}^{N}y_{i}^{k}$ denote the number of training examples that contain class $i$. Please note that labels for computing the class-specific embedding loss need to be processed into $\tilde y^{k}=\left[\tilde y_{1}^{k},...,\tilde y_{C}^{k} \right]=\left[2*y_{1}^{k}-1,...,2*y_{C}^{k}-1\right]\in {\left\{-1,1 \right\}}^{C}$, where $\left\{-1,1 \right\}$ indicates negative and positive.

\subsection{Approach Overview}
In order to make effective use of the linguistic modality in the long-tailed multi-label visual recognition task, we propose a novel framework (\textit{i.e.}, LMPT), as depicted in Fig.~\ref{fig:pipeline}. Text encoder from the pre-trained CLIP is used to encode the prompts and text descriptions (captions) of images. Only the parameters in the prompts are optimized, while the text encoder and image encoder are both kept frozen. We introduce two sorts of trainable prompts to obtain class embedding, which are jointly optimized by the classification loss $ \mathcal{L}_{cls}$ and class-specific embedding loss $ \mathcal{L}_{cse}$. Details of the aforementioned loss functions will be introduced in the later sections. 
\par

\subsection{Prompt Tuning}
Formally, the vision-language model consists of an image encoder $\boldsymbol{f}(\cdot)$ and a text encoder $\boldsymbol{g}(\cdot)$.
Following~\cite{zhou2022conditional}, a prompt is defined as:
\begin{equation}
o_{i}|_1^{M}=\left[\mathrm{V}\right]_{1}\left[\mathrm{V}\right]_{2}...\left[\mathrm{V}\right]_{m}...\left[\mathrm{V}\right]_{M}\left[\mathrm{CLASS}\right],
\end{equation}
where $i\in \left\{1,...,C\right\}$, $m\in \left\{1,...,M\right\}$, the $\left[\mathrm{CLASS}\right]$ token is replaced by the specific class name (\textit{e.g.}, “cat,” “dog”, “car”), each $\left[\mathrm{V}\right]_{m}$ is a learnable word embedding with the same dimension as normal word embeddings in the vocabulary (\textit{i.e.}, 512 for CLIP), and $M$ is a hyper-parameter specifying the number of context tokens. The prediction probability (classification output) $z$ is then computed as:
\begin{equation}
p(y=i \mid x)=\frac{\exp \left(\cos \left(\boldsymbol{g}\left(o_i\right), \boldsymbol{f}\left(x\right)\right) / \tau\right)}{\sum_{j=1}^C \exp \left(\cos \left(\boldsymbol{g}\left(o_j\right), \boldsymbol{f}\left(x\right)\right) / \tau\right)},
\end{equation}
where $\tau$ is a temperature parameter learned by CLIP
and $cos(\cdot, \cdot)$ represents cosine similarity.

\subsection{Class-Specific Embedding Loss}
\quad We introduce the class-specific embedding (CSE) loss to optimize the trainable fine-grained instance prompts by learning from text embeddings of captions. It tries to minimize the cosine distance of matching patches
and to increase the cosine distance of non-matching patches above the margin. Embedding loss is then computed as
\begin{equation}
\begin{aligned}
{\ell}_{ebd} & = \begin{cases}
                        \Delta_{i}^{k}, & \text{if}\quad \tilde y_{i}^{k} = 1, \\
                        \max \left(0, \mu-\Delta_{i}^{k}\right), & \text{if}\quad \tilde y_{i}^{k} = -1,
                        \end{cases} \\
& \Delta_{i}^{k} = 1-\cos\left(t_{i}^{k}, o_{i}|_{m}^{M}\right),
\end{aligned}
\end{equation}
where $\mu$ is the margin factor. Intuitively the embedding loss penalizes positive (\textit{i.e.}, prompts of matching classes) pairs that have large distances and negative (\textit{i.e.}, prompts of non-matching classes) pairs that have small distance (less than $\mu$).

\par

\begin{figure}
\begin{center}
\includegraphics[width=5cm]{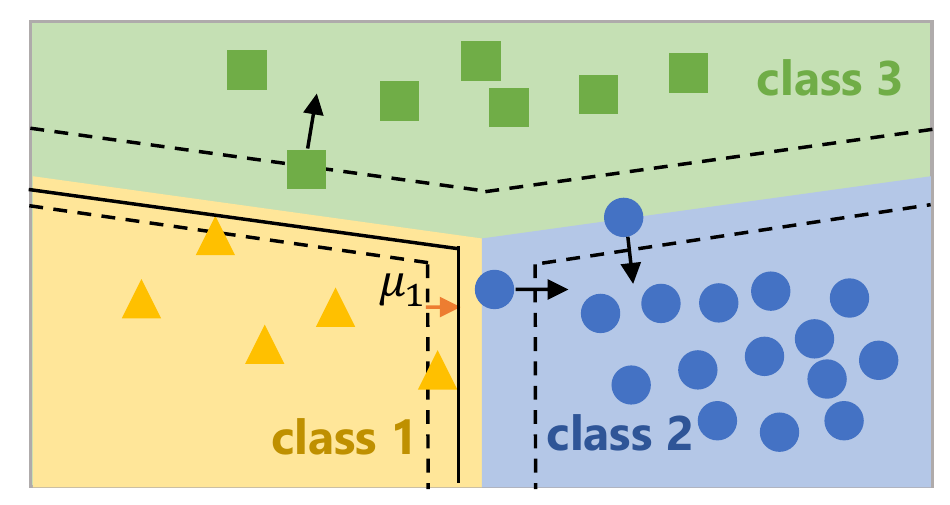}
\end{center}
\caption{The class margins (dotted lines) are enforced for generated samples by updating the decision boundary with respect to class margins.}
\label{fig:cseloss}
\vspace{-0.2em}
\end{figure}

LDAM~\cite{cao2019learning} has inspired the development of a decision boundary that is both robust and generalizable, capable of accurately classifying features that vary within a certain range. However, when applied to long-tailed datasets characterized by a significant class imbalance, models tend to exhibit greater sensitivity to more frequent classes. As a result, the performance of these models in less frequent classes is often poor. \par To address this issue, CSE loss employs the class-aware soft margin strategy to encourage the model to have the optimal trade-off between per-class margins by stimulating the minority classes to have larger margins, which can be viewed as regularization~\cite{wei2018margin}. More specifically, as illustrated in Fig.~\ref{fig:cseloss}, \textcolor[RGB]{44, 73, 200}{blue} samples (head classes) are classified incorrectly, and the model update gradient is shown with pointed arrows. \textcolor[RGB]{112, 173, 71}{Green} samples (medium classes) are classified correctly outside of the margin and the gradient is shown. Intuitively, the embedding loss does not give special consideration to the minority categories, but with the help of class-aware soft margin, the trade-off of $\mu_1$ (in Fig.~\ref{fig:cseloss}) can be optimized by shifting the decision boundary to encourage the tail classes to have larger margins. So \textcolor[RGB]{255, 204, 0}{yellow} samples (tail classes) are classified correctly outside of the original margin but within the enlarged margin, and the embedding loss has no gradient for these samples. Following the trade-off between the class margins, we adopt a class-aware margin for multiple classes of the form 
\begin{equation}
\label{eq:mu}
    \widetilde{\mu_{i}} \propto n_{i}^{-1/4} = \frac{\eta}{n_{i}^{1/4}}.
\end{equation}
Here $\eta$ is a hyper-parameter to be tuned. Therefore, when $y_{i}^{k} = -1$, the loss can be computed as $\max\left\{0, \widetilde{\mu_{i}}-\Delta_{i}^{k}\right\}$.

Meanwhile, our loss can be combined with a re-weighting strategy to be more efficient when it comes to long-tailed distribution data. We then define the reference weight based on the empirical class frequencies $\left\{n_1,...,n_C\right\}$ on the training set:
\begin{equation}
\label{eq:weight}
w_i = \frac{\left(1/n_i\right)^\gamma}{\sum_{i=1}^C \left(1/n_i\right)^\gamma},
\end{equation}
where $\gamma$ is a scale hyper-parameter to provide more flexibility. Hence, the re-weighted class-specific embedding loss is defined as:

\begin{equation}
\begin{aligned}
\label{eq:embeddingloss}
{\ell}_{cse} & =  
            \begin{cases}
            w_i\Delta_{i}^{k}, & \text{if}\quad \tilde y_{i}^{k} = 1, \\
            \max \left\{0, w_i\left(\widetilde{\mu_{i}}-\Delta_{i}^{k}\right)\right\}, & \text{if}\quad  \tilde y_{i}^{k} = -1,
            \end{cases} \\
\end{aligned}
\end{equation}

\begin{equation}
\label{eq:lcse}
\mathcal{L}_{cse} = \frac{\sum_{k=1}^{N}{\ell}_{cse}}{N}.
\end{equation}
\par

\begin{algorithm}
    \small
    \caption{Class-Specific Embedding Loss}
    \LinesNumbered 
    \label{ag:cseloss}
	\KwIn{Text embeddings of textual descriptions (captions) $t$, labels $\widetilde y$, prompt $o$}
	\KwOut{Class-Specific Embedding Loss $\mathcal{L}_{cse}$}
	\For{$k = 1,2,...,N$}{
		$\ell_{cse}=0$\;
        \For{$i=1,2,...,C$}{
        \textbf{Calculate} class-aware soft margin $\widetilde \mu_i$ by Eq.~\ref{eq:mu}\;
        \textbf{Calculate} weight $w_i$ by Eq.~\ref{eq:weight}\;
        \textbf{Calculate} $\Delta_i^k=1-\cos\left(t_{i}^{k}, o_{i}|_{m}^{M}\right)$\;
        \eIf{$\widetilde y_i^k=1$}{
			$\ell_{cse}=w_i\Delta_i^k$\;
		}{
			$\ell_{cse}=\text{ReLU}\left(w_i\left(\widetilde \mu_i-\Delta_i^k\right)\right)$\;
		}
      }
	}
    \textbf{Calculate} $\mathcal{L}_{cse}$ by Eq.~\ref{eq:lcse}.
    \vspace{-0.2em}
\end{algorithm}

The overall process of class-specific embedding loss is outlined in Algorithm~\ref{ag:cseloss}.

\subsection{Multi-Label Classification Loss}
\quad Our method can be easily combined with the existing multi-label classification loss functions~\cite{ridnik2021asymmetric,lin2017focal,cui2019class,wu2020distribution}, regardless of whether they are designed for long-tailed distributions or not. By blending the classification loss functions with our proposed CSE loss, our method facilitates prompt learning of more refined class descriptions and semantic relationships between categories, particularly between head and tail classes. \par
In this study, we introduce the distribution-balanced loss~\cite{wu2020distribution} as the classification loss function, which can be formulated as:
\begin{equation}
 r=\alpha+\sigma\left(\beta \times\left(\frac{\frac{1}{n_i}}{\sum_{i=1}^C\frac{1}{n_i}}-\theta\right)\right),   
\end{equation}
\begin{equation}
    v_i=-\kappa \times -\log \left(\frac{1}{n_i / N}-1\right),
\end{equation}
\begin{equation}
\begin{aligned}
{\ell}_{cls} & =  
            \begin{cases}
            -r\left(1-q_i^k\right)^\gamma\log\left(q_i^k\right), & \text{if}\quad y_{i}^{k} = 1,\\
            -\frac{r}{\zeta}\left(q_i^k\right)^\gamma\log\left(1-q_i^k\right), & \text{if}\quad y_{i}^{k} = -1,
            \end{cases} \\
\end{aligned}
\end{equation}
where $q_i^k=\sigma\left(z_i^k-v_i\right)$ is for positive instances, $q_i^k=\sigma\left(\zeta\left(z_i^k-v_i\right)\right)$ is for negative ones and $\alpha,\beta,\theta, \kappa,\zeta$ are hyperparameters. 
Then $\mathcal{L}_{cls}=\sum_{k=1}^{N}{\ell}_{cls}/N$. \par Hence, the overall training loss can be written as: 
\begin{equation}
\label{eq:overallloss}
\mathcal{L} = \lambda \mathcal{L}_{cls} + \left(1-\lambda \mathcal{L}_{cse} \right),
\end{equation}
where $\lambda \in [0,1]$ is a hyperparameter to balance $ \mathcal{L}_{cls}$ and $ \mathcal{L}_{cse}$.

\section{Experiment}

\begin{table*}[t]
\centering
\resizebox{\linewidth}{!}{
\begin{tabular}{l||c|c|c|c||c|c|c|c}
\hline
Datasets   & \multicolumn{4}{c||}{VOC-LT}     & \multicolumn{4}{c}{COCO-LT}    \\
\hline
Methods    & total & head  & medium & tail  & total & head  & medium & tail  \\
\hline
{\footnotesize \textup{\textbf{RN-50}}} &&&&&&&& \\
ERM & 70.86 & 68.91 & 80.20  & 65.31 & 41.27 & 48.48 & 49.06  & 24.25 \\
RW  & 74.70 & 67.58 & 82.81  & 73.96 & 42.27 & 48.62 & 45.80  & 32.02 \\
Focal Loss~\cite{lin2017focal} {\scriptsize \textup{ICCV'17}} & 73.88 & 69.41 & 81.43  & 71.56 & 49.46 & 49.80 & 54.77  & 42.14 \\
RS~\cite{shen2016relay} {\scriptsize \textup{ECCV'16}} & 75.38 & 70.95 & 82.94  & 73.05 & 46.97 & 47.58 & 50.55  & 41.70 \\
ML-GCN~\cite{chen2019multi} \scriptsize \textup{CVPR'19} & 68.92 & 70.14 & 76.41 & 62.39 & 44.24 & 44.04 & 48.36 & 38.96 \\
OLTR~\cite{liu2019large} \scriptsize \textup{CVPR'19} & 71.02 & 70.31 & 79.80 & 64.95 & 45.83 & 47.45 & 50.63 & 38.05 \\ 
LDAM~\cite{cao2019learning} \scriptsize \textup{NeurIPS'19} & 70.73 & 68.73 & 80.38 & 69.09 & 40.53 & 48.77 & 48.38 & 22.92 \\
CB Focal~\cite{cui2019class} \scriptsize \textup{CVPR'19} & 75.24 & 70.30  & 83.53  & 72.74 & 49.06 & 47.91 & 53.01  & 44.85 \\
BBN~\cite{zhou2020bbn} \scriptsize \textup{CVPR'20} & 73.37 & 71.31 & 81.76 & 68.62 & 50.00 & 49.79 & 53.99 & 44.91 \\
DB Focal~\cite{wu2020distribution} \scriptsize \textup{ECCV'20} & 78.94 & 73.22 & 84.18  & 79.30  & 53.55 & 51.13 & 57.05  & 51.06 \\
LTML~\cite{guo2021long} \scriptsize \textup{CVPR'21} & 81.44 & \textbf{75.68} & 85.53  & 82.69 & 56.90 & \textbf{54.13} & 60.59  & 54.47 \\
\hdashline[1pt/1pt]
CLIP~\cite{radford2021learning} \scriptsize \textup{ICML'21} & 84.30 & 63.60 & 88.03 & 97.03 & 56.19 & 35.73 & 60.52 & 68.45 \\
CoOp~\cite{zhou2022learning} \scriptsize \textup{IJCV'22} & 81.34 & 65.10  &	81.54	& 93.37 & 54.94 & 38.06 & 56.67 & 67.51  \\
CoCoOp~\cite{zhou2022conditional} \scriptsize \textup{CVPR'22} & 78.63 &	64.33 &	80.51 &	87.94 & 46.02 &	36.02 &	50.57 &	48.82  \\
DualCoOp~\cite{sun2022dualcoop} \scriptsize \textup{NeurIPS'22} & 81.03 & 66.45 & 80.53  & 92.33 & 53.11 & 40.48  & 55.20  & 62.11 \\
TaI-DPT~\cite{guo2023texts} \scriptsize \textup{CVPR'23} & 83.75 & 66.27 & 85.17  & 94.57 & 56.23 & 40.52  & 58.40  & 66.09 \\ \rowcolor[gray]{.9}
LMPT (ours) & 85.44 & 66.62	& 88.11	& 97.86 & 58.97 & 41.87 & 61.60 & 69.60 \\
\hline
{\footnotesize  \textup{\textbf{ViT-B/16}}} &&&&&&&& \\
CLIP~\cite{radford2021learning} \scriptsize \textup{ICML'21} & 85.77 & 66.52 & 88.93  & 97.83 & 60.17 & 38.52 & 65.06  & 72.28 \\
CoOp~\cite{zhou2022learning} \scriptsize \textup{IJCV'22} & 86.02 & 67.71 & 88.79 & 97.67 & 60.68 & 41.97 & 63.18 &	73.85 \\
CoCoOp~\cite{zhou2022conditional} \scriptsize \textup{CVPR'22} & 84.47 & 64.58 & 87.82 & 96.88 & 61.49 & 39.81 &	64.63 &	76.42 \\
\rowcolor[gray]{.9}
LMPT (ours)  & \textbf{87.88} & 72.10 & \textbf{89.26}  & \textbf{98.49} & \textbf{66.19} & 44.89  & \textbf{69.80}  & \textbf{79.08} \\
\hline
\end{tabular}
}
\caption {mAP performance of the proposed method and comparison methods. Above the dotted line is the performance of image-only models and below is that of vision-language models.}
\label{tab:wholetable}
\end{table*}

\subsection{Benchmark Setting}
Following \cite{wu2020distribution,guo2021long}, we conduct experiments on two datasets for long-tailed multi-label visual recognition: VOC-LT and COCO-LT~\cite{wu2020distribution}. They are artificially sampled from two multi-label recognition benchmarks, PascalVOC~\cite{everingham2015pascal} and MS-COCO~\cite{lin2014microsoft}, respectively. 

\subsection{Experimental Settings}
\textbf{Metrics.} As in \cite{liu2019large}, the classes are split into three groups by the number of their training examples: head classes each contain over 100 samples, medium classes each have between 20 and 100 samples, and tail classes with under
20 samples each. We use mean average precision (mAP) to evaluate the performance of long-tailed multi-label visual recognition for all the classes. 
\\
\textbf{Implementation Details.}  We adopt CLIP ResNet-50~\cite{he2016deep} or ViT-B/16~\cite{dosovitskiy2020image} as the visual encoder and use the corresponding CLIP Transformer as the text encoder. During training, the parameters of both the two encoders are kept frozen, and only learnable prompts are optimized. SGD optimizer is adopted to learn prompt tokens, and the training epochs are set to 30. The learning rates for COCO-LT and VOC-LT are empirically initialized with 1e-4, 5e-4, and decay by the cosine annealing rule during training. For loss functions, $\eta$ in Eq.~\ref{eq:mu}, $\gamma$ in Eq.~\ref{eq:weight} and $\lambda$ in Eq.~\ref{eq:overallloss} are set as 1.0, 1.0 and 0.5, respectively. Other hyperparameters in DB loss are set as the same as~\cite{wu2020distribution}. 


\subsection{Long-Tailed Multi-Label Visual Recognition} 
\quad To evaluate the effectiveness of the proposed method, firstly we compare it with previous methods of image-only models on the two long-tailed multi-label datasets. The compared methods include Empirical Risk Minimization (ERM), a smooth version of Re-Weighting (RW) using the inverse proportion to the square root of class frequency, Re-Sampling (RS)~\cite{shen2016relay}, Focal Loss~\cite{lin2017focal}, ML-GCN~\cite{chen2019multi}, OLTR~\cite{liu2019large}, LDAM~\cite{cao2019learning}, Class-Balanced (CB) Focal~\cite{cui2019class}, BBN~\cite{zhou2020bbn}, Distribution-Balanced (DB) Focal~\cite{wu2020distribution} and LTML~\cite{guo2021long}. The mAP performance of different methods is shown in Table~\ref{tab:wholetable}. The prior best performance is achieved by LTML – mAP of 81.44\% over all classes on VOC-LT and 56.90\% over all classes on COCO-LT. 
\par
Furthermore, we compare zero-shot and prompt learning methods based on CLIP on the two benchmarks. The mAP performance of these methods is shown in Table~\ref{tab:wholetable} as well. For a fair comparison, we initialize the prompt as the default hand-crafted one “a photo of a" for all the methods. The results show that when using ViT-B/16 as the backbone, even the overall mAP performance of zero-shot CLIP reaches 85.77\% and 60.17\%, which outperforms previous SOTA LTML by \textbf{4.33} points (85.77\% vs.81.44\%) and \textbf{3.27} points (60.17\% vs.56.90\%) on the two datasets, respectively. Therefore, it is meaningful to explore how to use prompt tuning based on CLIP effectively for better performance. From the perspective of prompt tuning methods, when using ResNet-50 as the backbone, the performance of our method on VOC-LT is more promising, which is \textbf{4.1} points, \textbf{6.81} points, \textbf{4.41} points and \textbf{1.69} points better than CoOp, CoCoOp, DualCoOp and TaI-DPT, which are popular prompt learning methods for single-label and multi-label recognition. The performance on COCO-LT is similar to that on VOC-LT, which is \textbf{4.03} points, \textbf{12.95} points, and \textbf{5.86} points better than CoOp, CoCoOp, and DualCoOp. When replacing the backbone with ViT-B/16, the overall mAP performance of our method can further boost up to 87.88\% and 66.19\%  on VOC-LT and COCO-LT, which is the current new \textbf{state-of-the-art} of the two datasets.

\begin{table}[t]
\footnotesize
\centering
\resizebox{\linewidth}{!}{
\begin{tabular}{l||c|c|c|c}
\hline
Datasets   & \multicolumn{4}{c}{VOC-LT}  \\
\hline
Methods    & total & head  & medium & tail \\
\hline
BCE & 82.18	& 64.90	& 83.17	& 94.30 \\
MLS & 84.30	& 64.31	& 84.82 & 97.47 \\
Focal Loss & 85.37 & 66.17 & 87.70 & 97.52 \\
CB Loss & 85.25	& 65.37	& 87.71	& 97.20 \\
R-BCE-Focal & 84.56 & 66.01 & 86.61 & 97.67 \\
ASL & 86.40	& 69.12	& 88.79	& 98.07 \\
DB Focal & 87.88 & 72.10 & 89.26 & 98.49 \\
\hline
\end{tabular}
}
\begin{tabular}{l||c|c|c|c}
\hline
Datasets   &  \multicolumn{4}{c}{COCO-LT}    \\
\hline
Methods    & total & head  & medium & tail  \\
\hline
BCE & 58.04	& 41.79	& 58.86	& 73.90 \\
MLS & 61.26 & 41.71 & 64.11 & 74.58 \\
Focal Loss & 54.40 & 37.60 & 59.36 & 62.33 \\
CB Loss & 56.45 & 34.61	& 58.77	& 74.52 \\
R-BCE-Focal & 60.13 & 38.11 & 64.87 & 72.79 \\
ASL & 64.89	& 43.18	& 68.22	& 78.43 \\
DB Focal & 66.19 & 44.89 & 69.80 & 79.08 \\
\hline
\end{tabular}
\caption {mAP performance of the proposed method with different multi-label loss functions.}
\label{tab:losstable}
\vspace{-0.2em}
\end{table}

\subsection{Ablation Analysis}
\begin{figure}[htbp]      
\centering
\begin{minipage}{0.49\linewidth}
        \centerline{\includegraphics[width=\textwidth]{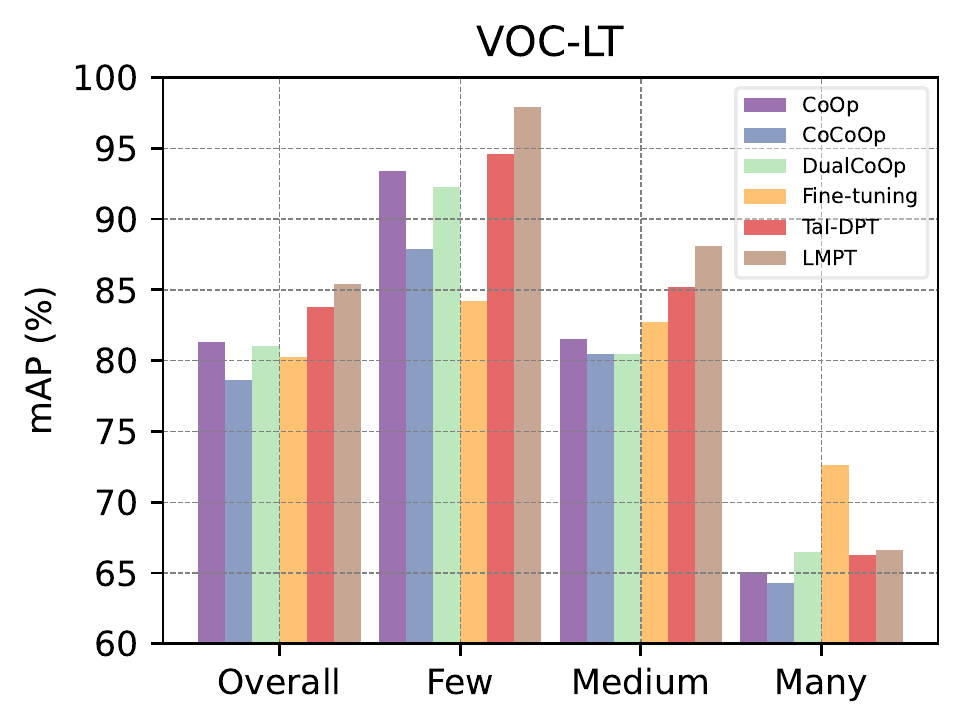}}
        \centerline{(a)}
	\end{minipage}
 \medskip
\begin{minipage}{0.49\linewidth}
        \centerline{\includegraphics[width=\textwidth]{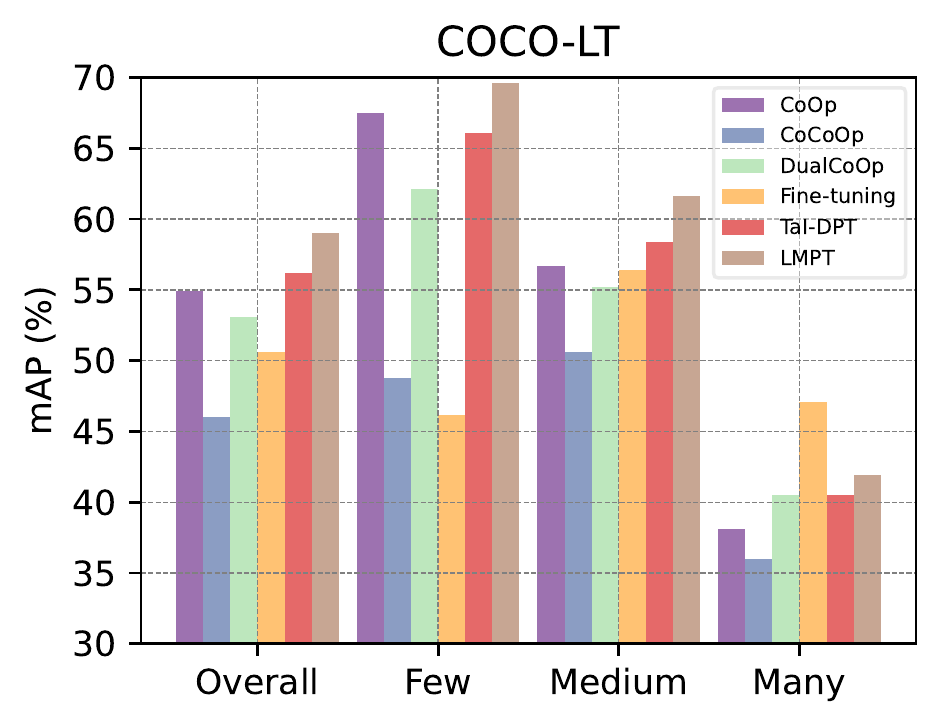}}
        \centerline{(b)}
	\end{minipage}
\caption{mAP performance of different methods w/o text supervision on two datasets.\;\;(a) VOC-LT. (b) COCO-LT.}      
\label{fig:dia}
\vspace{-1.5em}
\end{figure}

\begin{table*}[htbp]
\footnotesize
\centering
\resizebox{\linewidth}{!}{
\begin{tabular}{c|c|c|c||c|c|c|cc||c|c|c|cc}
\hline 
\multirow{2}{*}{\begin{tabular}[c]{@{}c@{}}Soft \\ Prompt\end{tabular}} & \multirow{2}{*}{\begin{tabular}[c]{@{}c@{}}Embedding \\ Loss\end{tabular}} & \multirow{2}{*}{\begin{tabular}[c]{@{}c@{}}Class-Aware \\ Soft Margin\end{tabular}} & \multirow{2}{*}{Re-weighting} & \multicolumn{4}{c}{VOC-LT} & \multirow{2}{*}{avg.$\Delta$}& \multicolumn{4}{c}{COCO-LT} & \multirow{2}{*}{avg.$\Delta$}\\ 
\cline{5-8}\cline{10-13} & & & & total & head & medium & tail & & total & head  & medium & tail& \\
\hline 
 &  &  &  & 85.77 & 66.52 & 88.93  & 97.83 & & 60.17 & 38.52 & 65.06  & 72.28 &\\
\ding{51} & & & & 86.02 & 67.71 & 88.79 & 97.67 &+0.29& 60.68 & 41.97 & 63.18 & 73.85& +0.91\\
\ding{51} & \ding{51} & & & 87.28 & 71.07 & 89.01 & 97.84 &+0.51& 65.34 & 44.27 & 69.39 & 77.96&+5.23\\
\ding{51} & \ding{51} & \ding{51} & & 87.62 & 72.01 & 89.26 & 98.13 &+1.99& 65.81 & 44.90 & 69.71 & 78.76&+5.79 \\
\ding{51} & \ding{51} & \ding{51} & \ding{51} & 87.88 & 72.10 & 89.26 & 98.49 &\textbf{+2.17}& 66.19 & 44.89 & 69.80 & 79.08 &\textbf{+5.98}\\
\hline
\end{tabular}
}
\caption {Ablation analysis on different components of the our method. “avg.$\Delta$" average performance improvement.} 
\label{tab:cse}
\vspace{-0.2em}
\end{table*}
\textbf{Components Analysis.} 
To further analyze which component makes our methods performant for LTML, we conduct a set of ablation studies and report the results in Table~\ref{tab:cse}. We first conduct experiments with CLIP and the mAP performances are 85.77\% on VOC-LT, 60.17\% on COCO-LT, which surprisingly outperforms the prior SOTA LTML. It indicates that pre-trained VLMs demonstrate a robust capability for visual recognition, providing a solid foundation for our approach. However, the mAP performance of the tail classes outperforms the head classes by nearly 30 points on both VOC-LT and COCO-LT. Then CoOp is benefited from soft prompts and the mAP performance is improved to 86.02\% on VOC-LT and 60.68\% on COCO-LT, with \textbf{0.25\%} and \textbf{0.51\%} increments. Besides, we design the class-specific embedding loss with class-aware soft margin and re-weighting to learn more fine-grained and class-related prompts that build semantic relationships across different classes, especially for the tail classes by encouraging those classes to have larger margins and weights. The mAP performances of head, medium, and tail classes after adding the embedding loss are all significantly improved and the overall mAP surpasses CoOp by \textbf{1.26\%} and \textbf{4.66\%} on VOC-LT and COCO-LT, which demonstrates our embedding loss can help prompts learn fine-grained classes descriptions and semantic relationships across the classes. Finally, the integration of CASM and RW strategy further improves the mAP performance slightly, mainly for the tail performance by \textbf{0.65\%} and \textbf{1.12\%} on VOC-LT and COCO-LT. \newline
\textbf{Multi-Label Classification Loss Functions.} We compare a number of multi-label classification loss functions, including Binary Cross-Entropy Loss (BCE), Multi-Label Soft Margin Loss (MSL), Focal Loss, CB Loss, R-BCE-Focal, Asymmetric Loss (ASL) and DB Focal. As illustrated in Table~\ref{tab:losstable}, DB Focal loss that takes the co-occurrence of labels and the dominance of negative labels into account works significantly better than other multi-label classification loss for the LTML task. \newline
\textbf{Effectiveness of Text Supervision.} We further compare our method with fine-tuning CLIP's image encoder when using ResNet-50 as the backbone to explore whether the significant effect of our approach is due to text supervision or simply because the CLIP's image encoder is so powerful. In order to prevent interference with the trained CLIP's image encoder during the fine-tuning phase, we only fine-tune a fully connected layer added at the end of the image encoder. The results are shown in Fig.~\ref{fig:dia}. Obviously, fine-tuning the image encoder shows promising results, but still largely underperforms LMPT, which suggests that the gradients that went through the text encoder provide more useful information.


\subsection{Case Analysis}
\begin{figure}[t]
\begin{center}
\includegraphics[width=8cm]{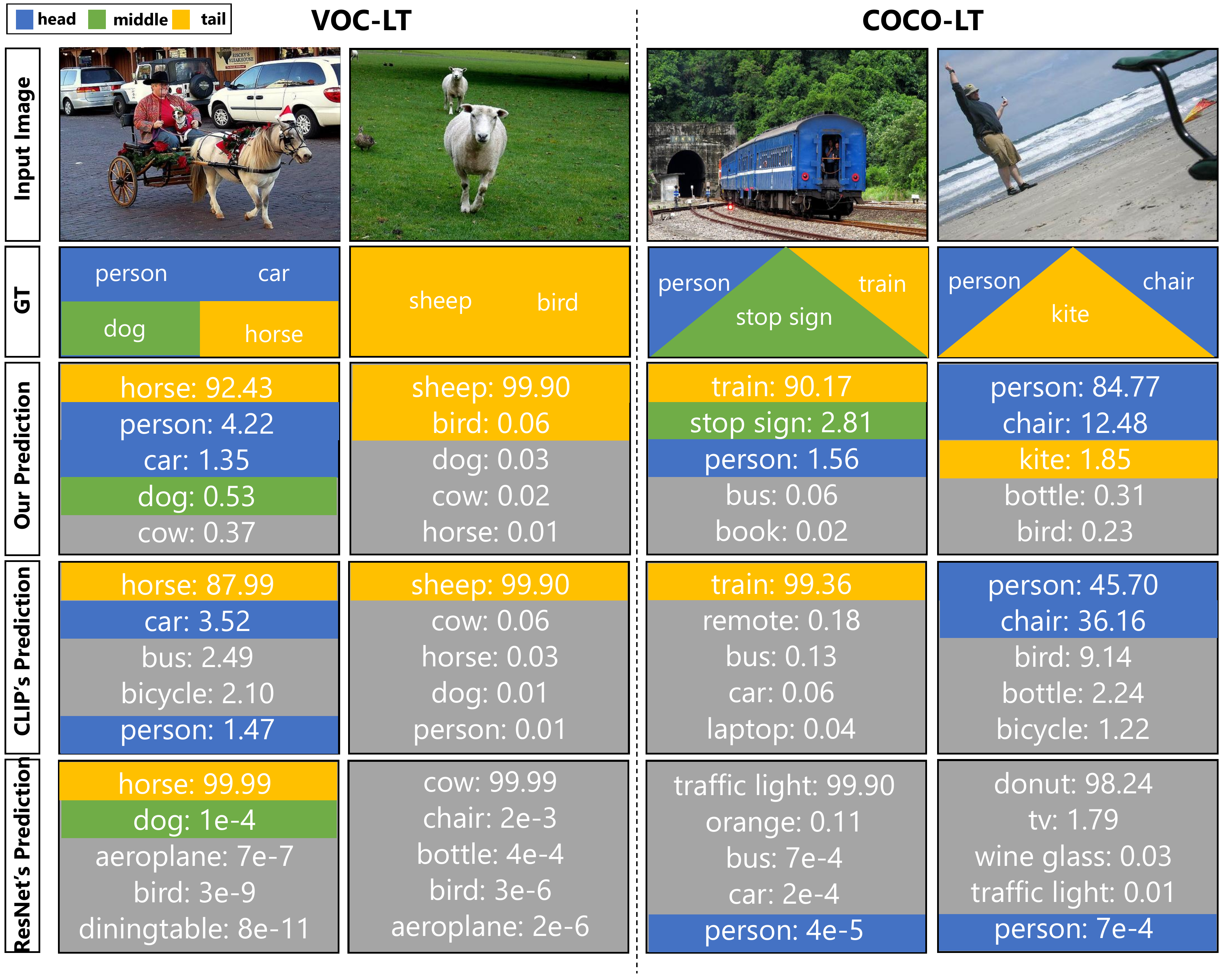}
\end{center}
\caption{Example decisions from our model, CLIP, and ResNet.}
\label{fig:example}
\end{figure}

To better understand how our method deals with long-tailed multi-label data, we performed qualitative experiments with ResNet, CLIP, and ours on COCO-LT and VOC-LT. Fig.~\ref{fig:example} shows several cases where the model justifies its abilities for the prediction. For example, in the third column, ResNet only recognizes $\left[\mathrm{person}\right]$ (belongs to head classes) and fails to classify the image to $\left[\mathrm{train}\right]$ (belongs to tail classes), which is a pervasive challenge encountered by image-only models. The emergence of CLIP is a great remedy for this issue, owing to its huge training data and effective text supervision. Nevertheless, simple hand-crafted templates as prompts still cannot accurately identify categories as they cannot describe the characteristics of each category. Understanding the inter-class relationships, particularly among head and tail categories, presents a formidable challenge in multi-label visual recognition, which is essential for achieving optimal performance in this domain. With the aid of our approach, utilizing prompts that learn from a large corpus of image-caption data, it has become feasible to discern the semantic relationships between categories and accurately predict the relevant categories of simple objects, even in challenging scenarios such as identifying $[\mathrm{stop}$ $\mathrm{sign}]$ from images. Therefore, our proposed method demonstrates significant advantages in effectively addressing the intricate relationship among multiple labels and the long-tailed problem with the aid of text supervision. 

\section{Conclusion}
In this work, we propose a new view of prompt tuning for long-tailed multi-label visual recognition by learning class-specific contexts from the alignment of prompts and textual description (caption), which complements more fine-grained features and builds semantic relationships across head and tail classes. Considering the class imbalance, a novel class-specific embedding loss with the class-aware soft margin and re-weighting strategy is introduced to promote increased generalization among the tail classes. Furthermore, we integrate a distribution-balanced loss as the classification loss function in consideration of its empirical efficacy compared to alternative loss functions. Our method exhibits significant improvement over the previous state-of-the-art (SOTA) and zero-shot CLIP on VOC-LT and COCO-LT. Additionally, 
We hope our approach will inspire future work in this field.

\bibliography{custom}
\bibliographystyle{acl_natbib}
\newpage
\appendix

\end{document}